\pdfoutput=1
\documentclass[11pt]{article}

\usepackage[]{ACL2024}

\usepackage{times}
\usepackage{latexsym}

\usepackage[T1]{fontenc}
\usepackage{booktabs}
\usepackage{graphicx}
\usepackage{colortbl}
\usepackage{multirow}
\usepackage{bbding} 
\usepackage{makecell}
\usepackage{bm} 
\usepackage{color} 
\usepackage{xcolor} 
\usepackage{textcomp,booktabs} 
\usepackage{amsmath} 
\usepackage{amsfonts}
\usepackage{amsthm,amsmath,amssymb}
\usepackage{mathrsfs}
\usepackage{subfigure}
\usepackage{mathtools}
\usepackage[utf8]{inputenc}

\usepackage{microtype}

\usepackage{inconsolata}
\usepackage{adjustbox}
\usepackage{CJKutf8}
\usepackage{makecell}
\usepackage{ulem}

\title{Alirector: Alignment-Enhanced Chinese Grammatical Error Corrector}

\author{
    Haihui Yang,
    Xiaojun Quan\thanks{\;\;Corresponding authors}\\
    School of Computer Science and Engineering, Sun Yat-sen University  \\
    \texttt{yanghh29@mail2.sysu.edu.cn, quanxj3@mail.sysu.edu.cn}
}

\begin{document}
\maketitle

\begin{abstract}
Chinese grammatical error correction (CGEC)  faces serious overcorrection challenges when employing autoregressive generative models such as sequence-to-sequence (Seq2Seq) models and decoder-only large language models (LLMs). While previous methods aim to address overcorrection in Seq2Seq models, they are difficult to adapt to decoder-only LLMs. In this paper, we propose an alignment-enhanced corrector for the overcorrection problem that applies to both Seq2Seq models and decoder-only LLMs. Our method first trains a correction model to generate an initial correction of the source sentence. Then, we combine the source sentence with the initial correction and feed it through an alignment model for another round of correction, aiming to enforce the alignment model to focus on potential overcorrection. Moreover, to enhance the model's ability to identify nuances, we further explore the reverse alignment of the source sentence and the initial correction. Finally, we transfer the alignment knowledge from two alignment models to the correction model, instructing it on how to avoid overcorrection. Experimental results on three CGEC datasets demonstrate the effectiveness of our approach in alleviating overcorrection and improving overall performance. Our code has been made publicly available\footnote{\url{https://github.com/yanghh2000/Alirector}}.

\end{abstract}

\section{Introduction}

\begin{figure}[t]
    \centering
    \includegraphics[width=\columnwidth]{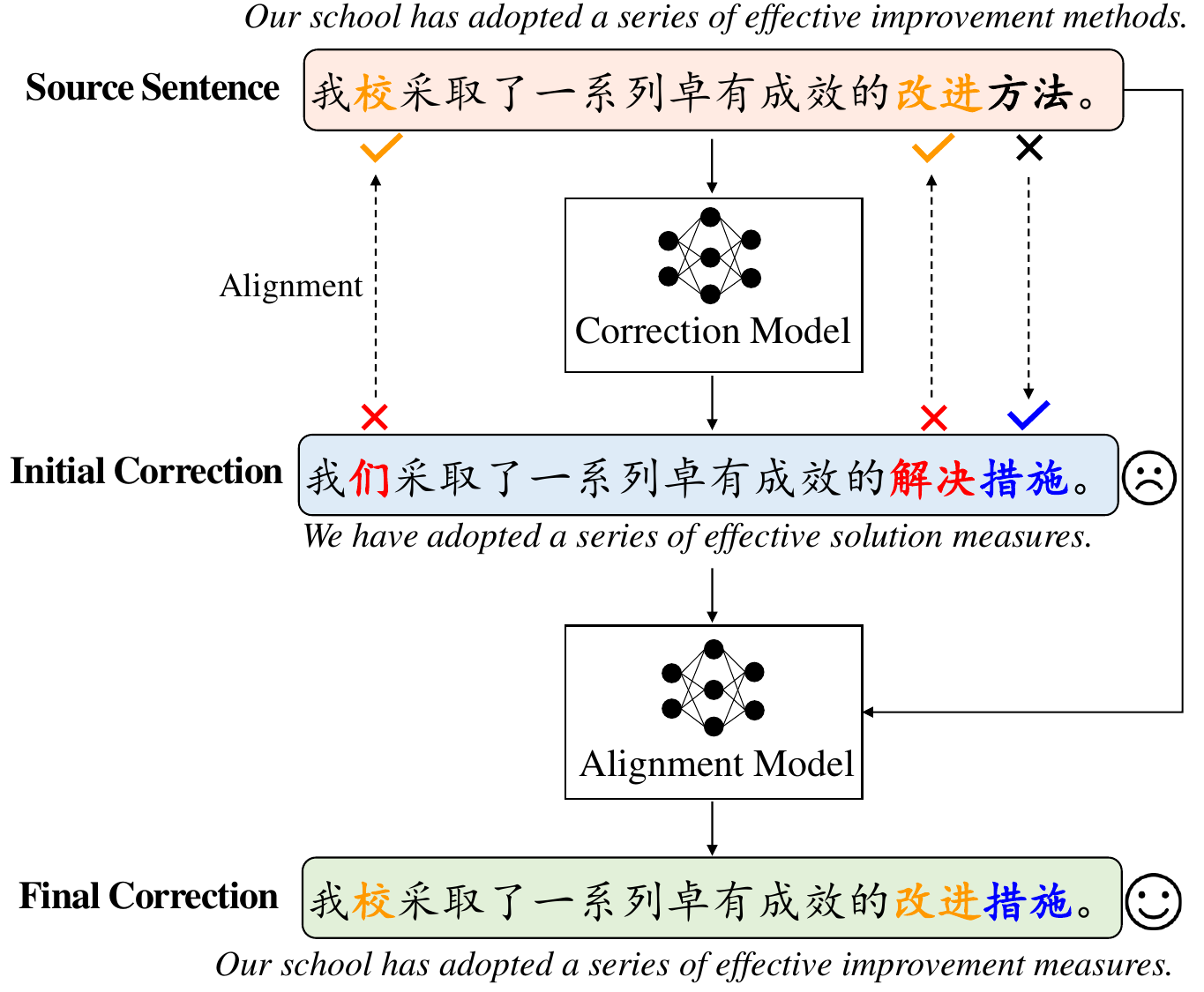}
    \caption{An illustration of addressing overcorrection through alignment of the source sentence and the initial correction. Overcorrected characters and their error-free counterparts are highlighted in red and orange, respectively. Correct edits are highlighted in blue.}
    \label{fig:example}
 \vspace{-5mm}
\end{figure}

Chinese grammatical error correction (CGEC) \citep{zhao2018overview}, which aims to identify and correct potential grammatical errors in given Chinese sentences while adhering to the principle of minimal editing, has broad applications in scenarios such as writing assistant and search engine \citep{wang2021comprehensive}. Chinese grammatical errors can be basically categorized into component missing, component redundancy, improper collocation, and improper word order \citep{ma-etal-2022-linguistic}, which are similar to those in English but tend to be more intricate due to the complexities of Chinese grammar.

Existing CGEC methods can be mainly divided into three categories: sequence-to-edit (Seq2Edit), sequence-to-sequence (Seq2Seq), and decoder-only large language models (LLMs). Seq2Edit methods treat CGEC as a sequence tagging task by predicting token-level edit operations \citep{liang-etal-2020-bert, zhang-etal-2022-mucgec}. While offering fast inference and robust error detection, these methods may compromise text fluency and exhibit weak migration ability due to the reliance on language-specific vocabulary \citep{s2a}. Seq2Seq methods tackle CGEC using neural machine translation techniques \citep{fu2018youdao, zhao2020maskgec} and excel in generating fluent sentences but often lack controllability. More recently, decoder-only LLMs have demonstrated breakthrough performance in various NLP tasks, showing significant potential in CGEC \citep{is_chatgpt, qu2023evaluating}. However, research suggests that decoder-only LLMs still fall short of surpassing lightweight state-of-the-art models \citep{zhang2023multi}.

Besides, Seq2Seq models and decoder-only LLMs may suffer from severe overcorrection issues, resulting in the modification of error-free characters of the source sentence \citep{park2020comparison}, as illustrated in Figure~\ref{fig:example}. This can be attributed to the tendency of these generative models to generate target sequences with higher probabilities and replace low-frequency words with more frequent ones \citep{s2a}. While increasing the number of training examples empirically alleviates this problem, obtaining high-quality annotated examples remains a challenge. Previous studies have explored mitigating overcorrection in Seq2Seq models. Among them, a copy module can be incorporated to enable the direct copying of correct tokens from source sentences to output sentences \citep{zhao-etal-2019-improving}. Another approach involves integrating error detection results from a Seq2Edit model into a Seq2Seq correction model \citep{li-etal-2023-templategec}. However, these methods prove challenging to migrate to decoder-only LLMs due to differences in their architectures. Given the emerging breakthroughs of LLMs in various NLP tasks,  there is an urgent need to explore their potential in CGEC, where the overcorrection problem presents a significant obstacle.

\begin{figure}[t]
    \centering
    \includegraphics[width=\columnwidth]{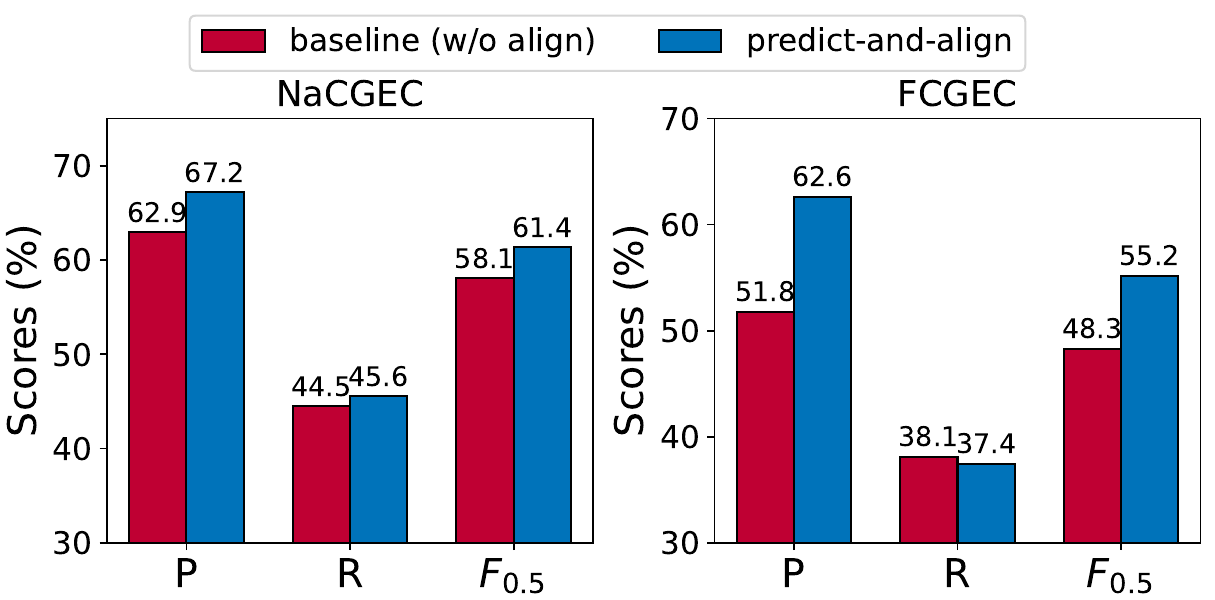}
    \caption{Preliminary results of predict-and-align on NaCGEC \citep{ma-etal-2022-linguistic} and FCGEC \citep{xu-etal-2022-fcgec} datasets with Baichuan2-7B model \citep{baichuan2}.}
    \label{fig:preliminary_result_baichuan}
 \vspace{-5mm}
\end{figure}

To fill this gap, we first explore a two-stage \textit{predict-and-align} method for mitigating overcorrection caused by Seq2Seq models and decoder-only LLMs. As illustrated in Figure~\ref{fig:example}, we first train a correction model to generate an initial correction of the source sentence. Then, we combine the source sentence with the initial correction and feed it through an alignment model for another round of correction. The alignment\footnote{For the explanation and discussion about the term "alignment" in this paper, please refer to Appendix~\ref{appendix:explanation_of_alignment}.} model is tasked not only with copying correct edits in the initial correction but also retaining error-free characters in the source sentence, thereby reducing overcorrections. Preliminary results in Figure~\ref{fig:preliminary_result_baichuan}  show that the two-stage method substantially enhances the overall performance of the original correction model.\footnote{More preliminary results are provided in Section~\ref{experiment:preliminary}.}

The above predict-and-align method requires deploying two models during inference, which is inefficient in terms of both time and storage. Therefore, we propose to enhance the correction model with knowledge acquired from the alignment model, resulting in an \textbf{ali}gnment-enhanced cor\textbf{rector} (Alirector) better at alleviating the overcorrection problem. Moreover, previous studies \citep{lu-etal-2022-fantastically, qin2023large} have shown that language models are sensitive to the ordering of the input sequence. Hence, we train another alignment model to explore the reverse combination of the source sentence and the initial correction. For knowledge transfer, we apply KL-divergence to constrain the output distributions of the correction model and the two alignment models, guiding the correction model on how to avoid overcorrection. Note that the proposed alignment method applies to both Seq2Seq models and decoder-only LLMs.

Extensive experiments were conducted on three CGEC datasets, and the experimental results demonstrate that our method achieves substantial improvements over baselines and effectively alleviates the overcorrection problem. Among various findings, our in-depth analysis reveals that the alignment information between the source sentence and the initial correction is crucial for mitigating overcorrection and improving the robustness of the correction model. Besides, we confirm that current decoder-only LLMs underperform Seq2Seq models, which warrants further investigation.

\section{Related Work}

\subsection{Traditional CGEC Methods}
Traditional CGEC methods typically follow the approaches used in English GEC, which are broadly categorized into Seq2Edit and Seq2Seq methods.

Seq2Edit methods \citep{awasthi-etal-2019-parallel, omelianchuk-etal-2020-gector, liang-etal-2020-bert, zhang-etal-2022-mucgec} treat GEC as a sequence editing task, which predicts token-level edit operations for the input sentence. PIE \citep{awasthi-etal-2019-parallel} utilizes BERT to iteratively predict edit labels. GECToR \citep{omelianchuk-etal-2020-gector} further extends the tag vocabulary with fine-grained edit tags.
\citet{liang-etal-2020-bert} and \citet{zhang-etal-2022-mucgec} explore adapting GECToR for CGEC tasks.
The strengths of Seq2Edit methods lie in its high inference efficiency and strong error detection performance. However, they rely heavily on manually designed vocabularies and language-specific lexical rules, limiting their adaptability \citep{s2a}.

On the other hand, Seq2Seq methods \citep{zhao-etal-2019-improving, zhao2020maskgec, kaneko-etal-2020-encoder, zhang-etal-2022-syngec} employ encoder-decoder models inspired by neural machine translation to model the GEC task, where the encoder encodes the source sentence and the decoder sequentially generates the target tokens. While \citet{kaneko-etal-2020-encoder} further adapts pre-trained knowledge into the encoder-decoder model, \citet{zhang-etal-2022-syngec} explore incorporating syntax information. 
Besides, efforts have been made to combine Seq2Edit and Seq2Seq to enhance the inference efficiency \citep{chen-etal-2020-improving-efficiency} or improve the correction results \citep{yuan-etal-2021-multi, s2a, li-etal-2023-templategec}. 

\subsection{LLMs for GEC}
With the success of LLMs across various NLP tasks, researchers have explored their potential for CGEC. Recent studies \citep{is_chatgpt, li2023effectiveness, qu2023evaluating, fan2023grammargpt} assess the performance of diverse LLMs, including both closed-source and open-source models, on the CGEC task. \citet{is_chatgpt} evaluate ChatGPT's performance on CGEC through in-context learning, highlighting its ability to generate fluent sentences as well as susceptibility to overcorrection. \citet{fan2023grammargpt} explore open-source LLMs for CGEC via instruction tuning \citep{sft}. \citet{zhang2023multi} suggest that fine-tuned LLMs still struggle to match the performance of existing state-of-the-art lightweight GEC models. Besides, some research endeavors \citep{kaneko2023controlled, song2023gee} aim at generating explanations for corrections utilizing LLMs' powerful capability. While these studies often overlook the overcorrection issue, our work presents a novel approach capable of mitigating overcorrection in LLMs.

\subsection{Overcorrection in GEC}
Seq2Seq models tend to generate sentences with higher probabilities and replace infrequent words with more frequent ones, leading to overcorrection. Previous works \citep{zhao-etal-2019-improving, s2a, li-etal-2023-templategec} expore various approaches to relieve this problem. \citet{zhao-etal-2019-improving} employ a copy module to directly copy the correct tokens from the source sentence to the output sentence. \citet{s2a} propose a sequence-to-action module based on the seq2seq model to generate a token-level action sequence. \citet{li-etal-2023-templategec} propose a two-stage approach by integrating detection results from a Seq2Edit model into a Seq2Seq correction model.
While these methods are challenging when applied to decoder-only LLMs due to architectural differences, the approach proposed in this work applies to both Seq2Seq models and decoder-only LLMs.

\section{Methodology}
\label{sec:method}
\vspace{-0.1cm}
\begin{figure*}[t]
    \centering
    \includegraphics[width=\textwidth]{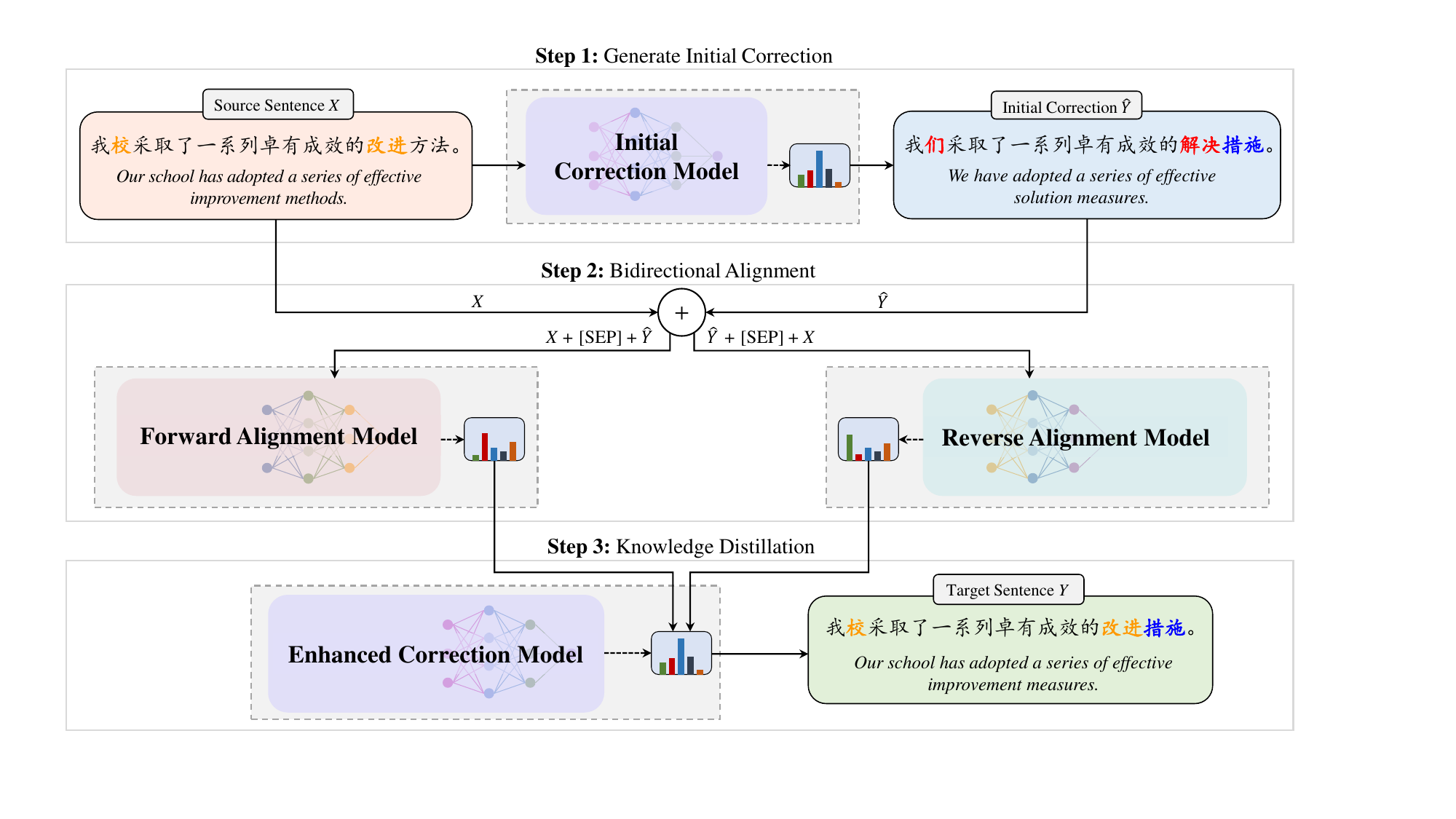}
    % \caption{An overview of our Alirector.}
    \caption{An overview of our proposed framework, which comprises three main steps. First, we train a correction model to generate an initial correction of the source sentence. Second, we perform bidirectional alignment by combining the source sentence with the initial correction forward and backward respectively, and passing each combination through an alignment model for another round of correction. Third, we employ knowledge distillation to transfer the knowledge from the two alignment models to the correction model.}
    \label{fig:overview}
    \vspace{-3mm}
\end{figure*}

As depicted in Figure~\ref{fig:overview}, our \textbf{ali}gnment-enhanced cor\textbf{rector} (Alirector) for Chinese grammatical error correction (CGEC) comprises three main steps to build. First, we train a correction model to generate an initial correction of the source sentence. Second, we perform bidirectional alignment by combining the source sentence with the initial correction forward and backward respectively, and passing each combination through an alignment model for another round of correction. Third, we employ knowledge distillation to transfer the knowledge from the two alignment models to the correction model. In the following sections, we first formulate the CGEC task and introduce the correction model in Section~\ref{method:gec}. Then, we delve into the details of the alignment models in Section~\ref{method:alignment_model} and specify the knowledge distillation in Section~\ref{method:kd}.
\vspace{-0.1cm}
\subsection{Correction Model}
\label{method:gec}
Given a source sentence $X =\{x_1,x_2,..,x_m\}$ that may contain grammatical errors, the goal of CGEC is to identify and correct the potential grammatical errors within $X$ and output the corresponding gold sentence $Y=\{y_1,y_2,..,y_n\}$. The models we investigate to implement the correction model include Transformer-based \citep{vaswani2017attention} Seq2Seq models and decoder-only LLMs.
\vspace{-0.1cm}
\paragraph{Seq2Seq}~The training objective of the Seq2Seq correction model is to minimize the negative log-likelihood (NLL) loss \citep{williams1989learning}:
\begin{equation}
\setlength\abovedisplayskip{0.2cm}
\setlength\belowdisplayskip{0.2cm}
    \mathcal{L}_{\text{gec}} = \sum\limits_{t=1}^{n}-\text{log}P(y_t|y_{<t}, X;\theta_1),
\end{equation}
where $y_{<t}$ represents the tokens preceding time step $t$, and $\theta_1$ denotes the trainable parameters.
\vspace{-0.1cm}
\paragraph{Decoder-only LLMs}
The input to the decoder-only correction model is formulated by converting $X$ and $Y$ into a natural language sequence $Z$ with an instruction template $\mathcal{T}_{\text{gec}}(X,Y)$:\footnote{Instruction templates are provided in Appendix~\ref{appendix:templates}.}
\begin{equation}
\setlength\abovedisplayskip{0.2cm}
\setlength\belowdisplayskip{0.2cm}
\begin{aligned}
    Z &= \mathcal{T}_{\text{gec}}(X,Y) \\
    &= \{\overbrace{z_{1},...,z_{i-1}}^{instruction},\overbrace{z_{i},...,z_{j}}^{X},\overbrace{z_{j+1},...,z_{j+n}}^{Y}\}.
\end{aligned}  
\label{eq:Y_GEC}
\end{equation}
Then, we compute the NLL loss on the target tokens as the training objective:
\begin{equation}
\setlength\abovedisplayskip{0.2cm}
\setlength\belowdisplayskip{0.2cm}
    \mathcal{L}_{\text{gec}} = \sum\limits_{t=j+1}^{j+n}-\text{log}P(z_{t}|z_{<t};\theta_{\text{2}}).
\end{equation}

\subsection{Alignment Model}
\label{method:alignment_model}
The purpose of our alignment model is to mitigate potential overcorrections in the initial correction generated by the aforementioned correction model. This is achieved by using a separate dataset and training the alignment model to predict the target sentence based on alignment information between the source sentence and the initial correction. Similar to the correction model, both Seq2Seq models and decoder-only LLMs can be employed to build the alignment model. However, to reduce the difficulty of transferring knowledge from the alignment model to the correction model, we require the two stages to share the same architecture\footnote{Knowledge distillation between different architectures may require an additional tokenizer and distribution alignment \citep{wan2024knowledge}.}.

\paragraph{Input Construction} 
We use $\hat{Y}$ to represent the initial output generated by the correction model for the source sentence $X$. Then, we construct the input to the alignment model based on $X$ and $\hat{Y}$ as follows. For Seq2Seq models, we simply concatenate $X$ and $\hat{Y}$ separated by ``[SEP]'' as the input, denoted as $X_\text{align}=X+[\text{SEP}]+\hat{Y}$. As for decoder-only LLMs, we follow Eq. (\ref{eq:Y_GEC}) and construct the input by transforming $X$, $\hat{Y}$ and $Y$ into a natural language sequence $W$ using another instruction template $\mathcal{T}_{\text{align}}$:
\begin{equation}
\setlength\abovedisplayskip{0.2cm}
\setlength\belowdisplayskip{0.2cm}
\begin{aligned}
   & W = \mathcal{T}_{\text{align}}(X,\hat{Y}, Y) \\
    &= \{...,\overbrace{w_{i},...}^{X},\overbrace{w_{j},...,w_{k}}^{\hat{Y}}, \overbrace{w_{k+1},...,w_{k+n}}^{Y}\}.
\end{aligned}  
\label{eq:align_input}
\end{equation}
\paragraph{Training Objective} 
The alignment model aims to predict $Y$ based on the alignment of $X$ and $\hat{Y}$. For Seq2Seq models, the training objective is:
\begin{equation}
\setlength\abovedisplayskip{0.2cm}
\setlength\belowdisplayskip{0.2cm}
    \mathcal{L}_{\text{align}} = \sum\limits_{t=1}^{n}-\text{log}P(y_t|y_{<t}, X_{\text{align}};\theta_{3}),
\end{equation}
where $\theta_3$ denotes the trainable parameters in the alignment model. 
As for decoder-only LLMs, the NLL loss is computed on the target tokens in $W$:
\begin{equation}
\setlength\abovedisplayskip{0.2cm}
\setlength\belowdisplayskip{0.2cm}
    \mathcal{L}_{\text{align}} = 
    \sum\limits_{t=k+1}^{k+n}-\text{log}P(w_t|w_{<t};\theta_4).
\end{equation}

\paragraph{Bidirectional Alignment}
Previous studies \citep{lu-etal-2022-fantastically, qin2023large} have shown that language models are sensitive to input ordering. Motivated by this, we further introduce bidirectional alignment by incorporating a reverse alignment model, which takes the combined source and initial correction in reverse order as input. For example, the input to the Seq2Seq-based reverse alignment model is $\hat{Y}+[\text{SEP}]+X$. 
Intuitively, the reverse alignment model may capture different information compared to the forward alignment model between the source sentence and the initial correction. Our empirical analysis in Section~\ref{analysis:ablation_study} also demonstrates that combining these two alignment models helps alleviate the impact of overcorrection and improves the overall robustness of the correction model.

\subsection{Bidirectional Alignment Distillation}
\label{method:kd}
The alignment models described above can be employed alongside the correction model in a two-stage predict-and-align paradigm to mitigate overcorrection. However, this approach presents two potential issues. Firstly, deploying both the correction model and the two alignment models during inference increases both time and storage requirements. Secondly, the correction model and the alignment models are trained separately, overlooking the possibility of mutual enhancement. To address these issues, we propose enhancing the correction model with knowledge distilled from the alignment models, guiding the correction model to avoid overcorrection, as well as eliminating the need for the alignment models during inference.

\paragraph{Knowledge Distillation}
We consider the two alignment models as the teachers and the correction model as the student for knowledge distillation \citep{hinton2015distilling}. During this process, only the parameters of the correction model are updated, while the parameters of the alignment models remain fixed. For training, we construct inputs for both the correction model and the alignment models following the methods introduced in Section \ref{method:gec} and Section \ref{method:alignment_model}, and obtain the final output logits over the target tokens. Let $z^c$, $z^f$, and $z^r$ denote the output logits from the correction model, the forward alignment model, and the reverse alignment model, respectively. We use KL-divergence as the distillation objective. The forward and reverse alignment distillation losses are defined as:
\begin{equation}
\setlength\abovedisplayskip{0.2cm}
\setlength\belowdisplayskip{0.2cm}
\begin{aligned}
    \mathcal{L}_{\text{kd}}^f=\mathcal{D}_{KL}(\sigma(\frac{z^f}{\tau})||\sigma(\frac{z^c}{\tau})) \\
    \mathcal{L}_{\text{kd}}^r=\mathcal{D}_{KL}(\sigma(\frac{z^r}{\tau})||\sigma(\frac{z^c}{\tau})),
\end{aligned}
\end{equation}
where $\tau$ is the temperature, $\sigma$ is the softmax function, and $\mathcal{D}_{KL}(\cdot)$ denotes the KL-divergence.

The overall distillation loss is the weighted sum of these two distillation losses:
\begin{equation}
\setlength\abovedisplayskip{0.2cm}
\setlength\belowdisplayskip{0.2cm}
\mathcal{L}_{\text{kd}}=\alpha\mathcal{L}_{\text{kd}}^{f}+(1-\alpha)\mathcal{L}_{\text{kd}}^{r},
\end{equation}
where $\alpha\in(0,1)$ is a hyperparameter.

\paragraph{Overall Objective}
To train the correction model, we formulate the overall objective by combining the GEC loss with the alignment distillation loss:
\begin{equation}
\setlength\abovedisplayskip{0.2cm}
\setlength\belowdisplayskip{0.2cm}
    \mathcal{L}=\mathcal{L}_{\text{gec}}+\beta\mathcal{L}_{\text{kd}},
\end{equation}
where $\beta$ is a hyperparameter that controls the importance of the distillation loss. More training details are provided in Appendix~\ref{appendix:training_details}.

\section{Experiments}

\begin{figure*}[t]
    \centering
    \includegraphics[width=0.95\textwidth]{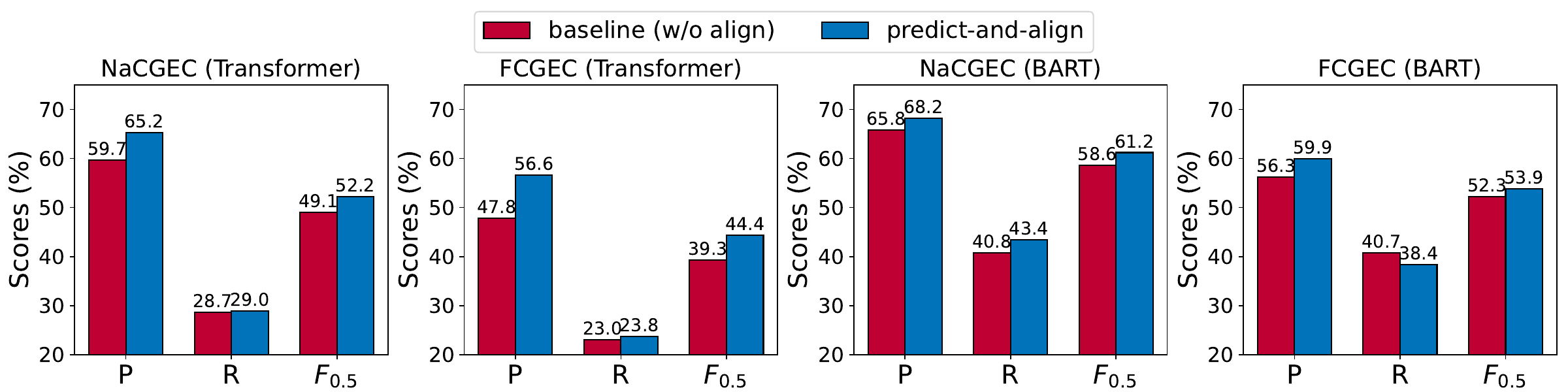}
    \caption{Preliminary results of predict-and-align on NaCGEC and FCGEC datasets with Transformer and BART.}
    \label{fig:preliminary_result_bart}
 \vspace{-2mm}
\end{figure*}

\subsection{Datasets and Metrics}
Based on the data sources, the datasets utilized in our experiments fall into two categories: i) datasets sourced from Chinese-as-a-Second-Language (CSL) learner texts, and ii) datasets sourced from Chinese native speaker texts. For CSL learner data, following previous works \citep{zhang-etal-2022-syngec, li-etal-2023-templategec}, we employ a combination of the Chinese Lang8 dataset \citep{zhao2018overview} and the HSK dataset \citep{zhang2009features} as our training set, MuCGEC-Dev \citep{zhang-etal-2022-mucgec} as the development set, and NLPCC18-Test \citep{zhao2018overview} as the test set. For native speaker data, we first randomly partition 1000 samples from the FCGEC \citep{xu-etal-2022-fcgec} training set as the development set, with the remainder used for training. For testing, we utilize FCGEC-Dev and NaCGEC-Test \citep{ma-etal-2022-linguistic} as our test sets\footnote{We use FCGEC-Dev here since we can not access the gold labels of  FCGEC-Test.}. Further details regarding the datasets can be found in Appendix~\ref{appendix:datasets}.

% Based on the data sources, the datasets utilized in our experiments fall into two categories: i) NLPCC18 \citep{zhao2018overview} sourced from Chinese-as-a-Second-Language (CSL) learners, and ii) FCGEC \citep{xu-etal-2022-fcgec} and NaCGEC \citep{ma-etal-2022-linguistic} from Chinese native speakers. For CSL learner data, following \citet{zhang-etal-2022-syngec}, we employ a combination of the Chinese Lang8 dataset \citep{zhao2018overview} and the HSK dataset \citep{zhang2009features} as our training set, MuCGEC-Dev \citep{zhang-etal-2022-mucgec} as the development set, and NLPCC18-Test as the test set. Regarding the native speaker data, we first randomly partition 1000 samples from the FCGEC training set for the development set, with the remainder used for training. Additionally, we utilize FCGEC-Dev and NaCGEC-Test as our test sets. Further details regarding the datasets can be found in Appendix~\ref{appendix:datasets}.

For evaluation metrics, we follow previous work and report word-level \emph{precision} (P)/\emph{recall} (R)/\emph{F-measure} ($F_{0.5})$ results on NLPCC18-Test using the official MaxMatch scorer \citep{dahlmeier-ng-2012-better} and PKUNLP word segmentation tool provided by \citet{zhao2018overview}.
For FCGEC-Dev and NaCGEC-Test, we report the character-level P/R/$F_{0.5}$ scores using the ChERRANT scorer\footnote{\url{https://github.com/HillZhang1999/MuCGEC/tree/main/scorers/ChERRANT}}.

\subsection{Base Models and Baselines}
As previously mentioned, the proposed method applies to both Seq2Seq models and decoder-only LLMs. For Seq2Seq, we choose Transformer-large and Chinese BART-large \citep{shao2021cpt} as the base models. For decoder-only LLMs, we choose Baichuan2-7B \citep{baichuan2}, a powerful Chinese LLM, and Chinese-LLaMA2-7B\footnote{\url{https://huggingface.co/Linly-AI/Chinese-LLaMA-2-7B-hf}}, which is obtained by incremental training of LLaMA2 \citep{llama2} with Chinese corpus. 

For comparison, we first employ the following Seq2Seq models as baselines. \textbf{Vanilla Fine-tuning (FT)} means directly fine-tuning the base models on the entire training set. \textbf{TemplateGEC} \citep{li-etal-2023-templategec} constructs a detection template to integrate the Seq2Edit and Seq2Seq methods.~\textbf{SynGEC} \citep{zhang-etal-2022-syngec} incorporates syntax information into Seq2Seq models. \textbf{Copy} \citep{zhao-etal-2019-improving} employs a copy mechanism for Seq2Seq models to directly copy unchanged words from the source sentence to the target sentence. 
Besides, we also employ decoder-only baselines. Except for vanilla fine-tuning, we implement the copy method \citep{zhao-etal-2019-improving} in decoder-only LLMs for comparison.
The implementation details and hyperparameter settings are presented in Appendix~\ref{appendix:implementation_details}.

\subsection{Preliminary Results}
\label{experiment:preliminary}
As mentioned earlier, we conducted preliminary experiments of the predict-and-align method on NaCGEC and FCGEC datasets. In addition to the results shown in Figure~\ref{fig:preliminary_result_baichuan} using Baichuan2-7B, we also present the results with two Seq2Seq models, namely Transformer-large and BART-large, in Figure~\ref{fig:preliminary_result_bart}.
From these results, we observe that after alignment, both Baichuan2-7B and Transformer exhibit a substantial performance improvement, especially in precision, revealing the potential of alignment in enhancing overall performance and mitigating overcorrection. While BART's performance improvement may not be as remarkable as Baichuan2, the alignment approach still demonstrates favorable enhancement. More analysis and discussion regarding the potential of the alignment method are presented in Appendix~\ref{appendix:alignment_model}.

\subsection{Main Results}
\label{experiment:main_results}
The main results are shown in Table~\ref{tab:main_results}.\footnote{More experimental results can be found in Appedix~\ref{appendix:more_exp_results}.}~We note that our Alirector consistently outperforms all baselines in $F_{0.5}$ across all the datasets, demonstrating the effectiveness of this method. In contrast, the Copy method even underperforms vanilla fine-tuning in some cases. Besides, Alirector achieves considerable improvements in precision across all the datasets, highlighting the efficacy of this approach in mitigating overcorrection. 
Further analysis regarding the effect of Alirector on reducing overcorrection is presented in Section~\ref{analysis:overcorrection}.
\definecolor{mygray}{gray}{.9}
\definecolor{ggreen}{rgb}{0.0, 0.6, 0.0}

\begin{table*}[t]
    \begin{adjustbox}{width=0.95\textwidth,center}
        \begin{tabular}{l c ccc ccc ccc ccc ccc}
            \toprule
            \multirow{2}{*}{\textbf{Model}}      & \multirow{2}{*}{\textbf{Method}}           & \multicolumn{3}{c}{\textbf{NLPCC18-Test}} & \multicolumn{3}{c}{\textbf{NaCGEC-Test}}       & \multicolumn{3}{c}{\textbf{FCGEC-Dev}}   \\
                &        & \textbf{P} & \textbf{R} & $\textbf{F}_{0.5}$       & \textbf{P} & \textbf{R} & $\textbf{F}_{0.5}$    & \textbf{P} & \textbf{R} & $\textbf{F}_{0.5}$ \\ 
            \hline
            \\[-0.9em]
             \multirow{5}{*}{Transformer}       & Vanilla FT       & 42.37  & 23.49  & 36.50   & 59.67  & 28.69  & 49.07       & 47.83  & 22.99  & 39.33        \\
                & TemplateGEC    & 42.00  & 22.20  & 35.60      & -  & -  & -    & -  & -  & -     \\
                & SynGEC        & 41.44  & \textbf{28.28}  & 37.91       & 51.45  & \textbf{39.69}  & 48.57       & 38.00  & \textbf{32.18}  & 36.67   \\
                & Copy       & 43.16  & 23.58  & 37.01     & 64.61  & 26.42  & 50.12      & 48.95  & 19.77  & 37.79  \\
                \rowcolor{mygray}  \cellcolor{white}  &  Alirector & \textbf{45.98} & 22.87 & \textbf{38.25} & \textbf{65.44} & 31.27 & \textbf{53.70} & \textbf{57.86} & 24.15 & \textbf{45.23} \\
            \hline 
            \\[-0.9em]
             \multirow{5}{*}{BART}       & Vanilla FT       & 50.63  & 31.83  & 45.28     & 65.85  & 40.79  & 58.64       & 56.26  & \textbf{40.71}  & 52.27   \\
                & TemplateGEC     & \textbf{54.50}  & 27.40  & 45.50      & -  & -  & -     & -  & -  & -    \\
                & SynGEC         & 49.96  & 33.04  & 45.32       & 63.76  & \textbf{47.41}  & 59.65     & 53.11  & 39.45  & 49.67   \\
                & Copy      & 51.25  & 32.55  & 45.97    & 66.67  & 41.88  & 59.61       & 58.55  & 38.46  & 53.01  \\
                \rowcolor{mygray}  \cellcolor{white}  &  Alirector      & 51.76  & \textbf{33.49}  & \textbf{46.67}    & \textbf{68.11}  & 43.87  & \textbf{61.33}         & \textbf{58.78}  & 39.15  & \textbf{53.42}   \\
            \hline 
            \\[-0.9em]
             \multirow{3}{*}{Baichuan2-7B}     & Vanilla FT    & 51.69  & 27.92  & 44.17    & 62.93   & 44.50  & 58.12       & 51.77  & 38.10  & 48.31         \\
                & Copy         & 51.56  & \textbf{28.53}  & 44.39   & 62.27  & 44.20  & 57.56      & 53.47  & 35.51  & 48.56          \\
                \rowcolor{mygray}  \cellcolor{white}  &  Alirector      & \textbf{52.27}   & 27.14  & \textbf{45.01}       & \textbf{66.04}  & \textbf{45.91}  & \textbf{60.71}         & \textbf{58.55}  & \textbf{39.74}  & \textbf{53.49}   \\
            \hline 
            \\[-0.9em]
             \multirow{3}{*}{Chinese-LLaMA2-7B}       & Vanilla FT    & 45.85  & 27.44  & 40.43   & 61.93  & 30.31   & 51.24        & 50.15  & 26.19  & 42.39          \\
                & Copy       & 46.53  & \textbf{27.93}  & 41.06       & 62.15  & 30.54  & 51.49      & 48.04  & 28.35   & 42.18           \\
                \rowcolor{mygray}  \cellcolor{white}  &  Alirector      & \textbf{47.43}  & 26.96  & \textbf{41.18}      & \textbf{62.60}  & \textbf{32.90}  & \textbf{53.03}      & \textbf{52.64}  & \textbf{28.47}   & \textbf{45.00}         \\
            \bottomrule
        \end{tabular}
    \end{adjustbox}
    \caption{Overall results on NLPCC18-Test, NaCGEC-Test, and FCGEC-Dev datasets. The results of TemplateGEC \citep{li-etal-2023-templategec} and SynGEC \citep{zhang-etal-2022-syngec} on NLPCC18 are cited from the original papers, and other results including Copy \citep{zhao-etal-2019-improving} are implemented by us. Best results are highlighted in bold.}
    
    \label{tab:main_results}
    \vspace{-3mm}
\end{table*}

Moreover, we make several interesting observations.
First, despite the notable enhancement achieved by Alirector, the decoder-only LLMs of Baichuan2 and Chinese-LLaMA2 still struggle to outperform BART. This can be attributed to the fact that BART's pre-training involves a series of denoising tasks utilizing strategies like token masking, token deletion and text infilling, which are naturally suitable for the CGEC/GEC task \citep{lewis-etal-2020-bart, wang2023chinese}. 
Second, different models exhibit varying degrees of improvement by employing Alirector, with decoder-only LLMs generally experiencing more notable improvements than Seq2Seq models.~The performance of Chinese-LLaMA2 is much worse than Baichuan2, which may be attributed to their different capabilities achieved through pre-training.
Third, Alirector yields more pronounced improvements on the FCGEC and NaCGEC datasets than on the NLPCC18 dataset. We attribute this discrepancy to the differing error distributions across datasets. The errors in NLPCC18, derived from Chinese-as-a-Second-Language learners, are less common, while the errors in FCGEC and NaCGEC, stemming from native speakers, exhibit more prevalent patterns that are easier for the model to learn.

\section{Analysis}
\begin{figure}
	\centering
\includegraphics[width=0.72\columnwidth]{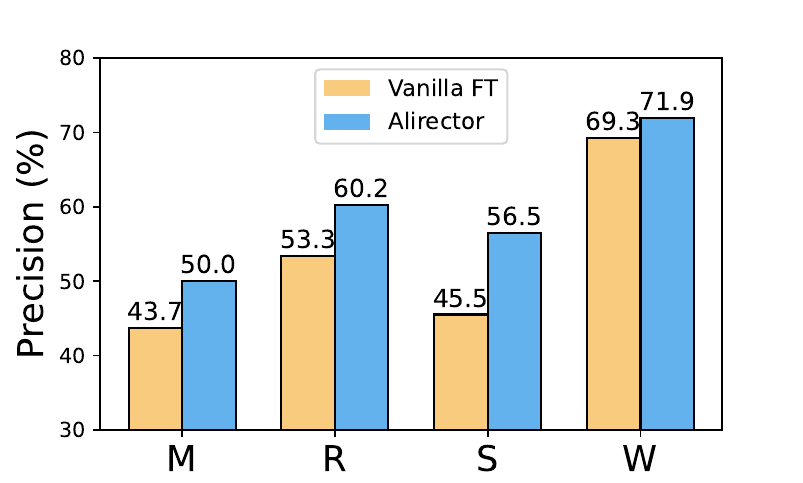}
	\caption{Results of precision for different error types, including \emph{missing} (M), \emph{redundant} (R), \emph{substitution} (S), and \emph{word-order} (W), on the FCGEC-Dev test set.}
	\label{fig:error_types}
 \vspace{-2mm}
\end{figure}

\subsection{Overcorrection Mitigation}
\label{analysis:overcorrection}
To further verify the effectiveness of Alirector in mitigating overcorrection, we use Baichuan2-7B as the backbone and present fine-grained precision results across the four categories of CGEC errors, including \emph{missing} (M), \emph{redundant} (R), \emph{substitution} (S), and \emph{word-order} (W), in Figure~\ref{fig:error_types}. Moreover, we present in Table~\ref{tab:error_types} the number of overcorrections and undercorrections that Alirector reduces on the four error types compared to Baichuan2-7B. The results depicted in Figure~\ref{fig:error_types} and summarized in Table~\ref{tab:error_types} demonstrate that Alirector significantly enhances precision for all error types while notably decreasing the number of overcorrections without deteriorating undercorrection, particularly for the \emph{redundant} and \emph{substitution} types. These findings support the effectiveness of Alirector in mitigating overcorrection induced by generative language models and in enhancing the robustness of our method across different error types. For a more intuitive illustration of Alirector's effectiveness, we provide a case study in Appendix~\ref{appendix:case_study}.

\begin{table}[t]

    \begin{adjustbox}{width=0.75\width,center}
    \begin{tabular}{c|ccc}
        \toprule
        \multirow{2}{*}{\textbf{Type}} & \multicolumn{3}{c}{ {\#Overcorrections} / {\#Undercorrections}} \\
        \cline{2-4}
        & & \textbf{Vanilla FT}  & \textbf{Alirector}  \\
        \hline 
        \textbf{M}  & & 129 / 259  & 113 (-12.4\%) / 245 \\
        \textbf{R}  & & 203 / 181  & 152 (-25.1\%) / 183 \\
        \textbf{S}  & & 91  / 226  & 67 (-26.4\%)  / 215 \\
        \textbf{W}  & & 39 / 140   & 34 (-12.8\%)  / 141 \\
        \hline
        \textbf{All}  & & 462 / 806 & 366 (-20.8\%) / 784  \\
        \bottomrule
    \end{tabular}
    \end{adjustbox}
    \caption{The number of overcorrections and undercorrections reduced by Alirector over direct fine-tuning for different error types on FCGEC-Dev.}
    
    \label{tab:error_types}
\vspace{-3mm}
\end{table}

\subsection{Ablation Study}
\label{analysis:ablation_study}
To investigate the contribution of key components of our approach, we conduct in-depth ablation experiments on NaCGEC-Test and FCGEC-Dev datasets using BART and Baichuan2-7B.

\paragraph{Distillation from Alignment Models}
We first ablate different alignment distillation components in turn to analyze their contribution. As shown in Table~\ref{tab:ablation_study}, while removing either forward distillation $\mathcal{L}_{\text{kd}}^{f}$ or reverse distillation $\mathcal{L}_{\text{kd}}^{r}$ causes noticeable performance degradation, there is a significant performance drop after removing the overall distillation loss $\mathcal{L}_{\text{kd}}$, particularly in recall and $F_{0.5}$. This indicates that bidirectional alignment contributes more to performance improvement through knowledge distillation compared to unidirectional alignment. Moreover, a notable drop in precision is observed when removing the entire distillation loss (as shown in the line labeled \textit{w/o} $\mathcal{L}_{\text{kd}}$), suggesting that the alignment distillation is essential for our method to mitigate overcorrection.

\begin{table}[t]
    \centering
    \resizebox{\columnwidth}{!}{
    \begin{tabular}{l|ccc ccc}
        \toprule
        \multirow{2}{*}{\textbf{Method}} & \multicolumn{3}{c}{\textbf{NaCGEC-Test}}  & \multicolumn{3}{c}{\textbf{FCGEC-Dev}}  \\
        & \textbf{P} & \textbf{R} & $\textbf{F}_{0.5}$    & \textbf{P} & \textbf{R} & $\textbf{F}_{0.5}$ \\ 
        \hline 
        \multicolumn{7}{c}{\textbf{BART}} \\
        \hline 
        \textbf{Alirector}      & 68.11&\textbf{43.87}&\textbf{61.33}       & 58.78&\textbf{39.15}&\textbf{53.42}   \\
        \ \textit{\, w/o} $\mathcal{L}_{\text{kd}}^{f}$      & \textbf{68.30}&40.44&60.03       & \textbf{59.70}&36.46&52.95   \\
        \ \textit{\, w/o} $\mathcal{L}_{\text{kd}}^{r}$      & 68.19&43.41&61.21       & 59.56&37.41&53.25   \\
        \ \textit{\, w/o} $\mathcal{L}_{\text{kd}}$      & 67.17&40.79&59.48       & 56.26&40.71&52.27   \\
        \hline 
        \textit{disc.} $\text{source}$      & 65.44 & 41.87 & 58.82        & 57.62 & 38.69 & 52.48   \\
        \textit{disc.} $\text{predict}$      & 67.93  & 39.53  & 59.40       & 59.22  & 35.08 & 52.05   \\
        \hline 
        \hline 
        \multicolumn{7}{c}{\textbf{Baichuan2-7B}} \\
        \hline 
        \textbf{Alirector}      & 66.04&\textbf{45.91}&\textbf{60.71}       & \textbf{58.55}&\textbf{39.74}&\textbf{53.49}   \\
        \ \textit{\, w/o} $\mathcal{L}_{\text{kd}}^{f}$      & 65.92&43.72&59.84       & 57.88&38.57&52.62  \\
        \ \textit{\, w/o} $\mathcal{L}_{\text{kd}}^{r}$      & \textbf{66.91}&40.99&59.40       & 55.99&36.66&50.65   \\
        \ \textit{\, w/o} $\mathcal{L}_{\text{kd}}$      & 62.93&44.50&58.12       & 51.77&38.10&48.31   \\
        \hline 
        \textit{disc.} $\text{source}$      & 59.98 & 49.46 & 57.53       & 51.46 & 39.22 & 48.44   \\
        \textit{disc.} $\text{predict}$      & 66.05 & 41.78 & 59.18       & 53.47 & 35.51 & 48.56   \\
        \bottomrule
    \end{tabular}
    }
    \caption{Results of ablation study on NaCGEC-Test and FCGEC-Dev, where ``\textit{disc.}'' is short for ``discard''. }
    
    \label{tab:ablation_study}
\vspace{-4mm}
\end{table}

\paragraph{Input of Alignment Models}
To further investigate the effect of the alignment between the source sentence and the initial correction, we conduct additional experiments by ablating the source sentence or initial correction from the input of the alignment models during training Alirector. To keep the format of the input, we ablate the source sentence by replacing it with the initial correction, e.g., $\hat{Y}+[\text{SEP}]+\hat{Y}$ for Seq2Seq. Similarly, we construct the input as $X+[\text{SEP}]+X$ when ablating the initial correction. As shown in Table~\ref{tab:ablation_study}, we observe that ablating the source sentence causes an obvious decline in precision while ablating the initial correction leads to a notable drop in recall. These findings highlight the role of alignment in reducing both overcorrection and undercorrection.

\subsection{Impact of $\alpha$ and $\beta$}
\begin{figure}
	\centering
        \includegraphics[width=0.82\columnwidth]{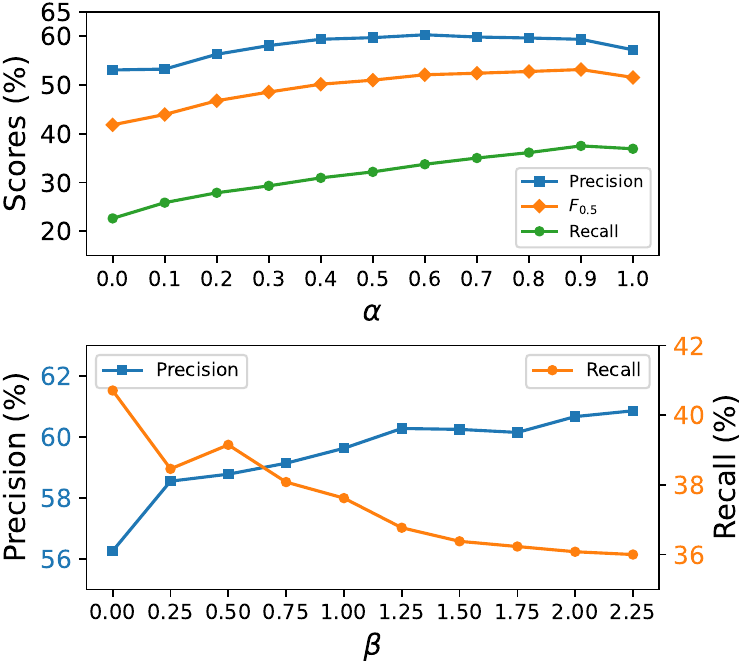}
	\caption{Results of our method on FCGEC development set with different values of $\alpha$ and $\beta$ that control the weight of forward and reverse alignment losses.}
	\label{fig:alpha_beta}
 \vspace{-4mm}
\end{figure}

The training objective of Alirector involves $\alpha$ to control the weight of forward and reverse alignment losses, as well as $\beta$ to balance between the original GEC loss and the distillation loss. To investigate their impact on model performance, we use BART as the backbone and show the results of different values of $\alpha$ and $\beta$ on the FCGEC development set in Figure~\ref{fig:alpha_beta}, where we change one while fixing the other. From the first subfigure, we observe that as $\alpha$ increases, the P/R/$F_{0.5}$ scores consistently rise and achieve the best results around 0.9. The second subfigure shows that as $\beta$ increases, the precision rises accordingly while recall gradually falls. This trend indicates that $\beta$ plays a role in balancing between precision and recall. Similar trends can be observed when other models are employed, and the optimal values of $\alpha$ and $\beta$ are provided in Table~\ref{tab:hyperparameter}.

\section{Conclusion}
In this paper, we first investigate a predict-and-align method that effectively leverages alignment information between the source sentence and the initial correction to alleviate the overcorrection issue in CGEC. Then, we propose transferring knowledge from the alignment process to enhance the correction model, resulting in an improved model termed Alirector. Experimental results on three CGEC datasets showcase the efficacy of our approach in mitigating overcorrection for both Seq2Seq models and decoder-only LLMs. Detailed analysis further demonstrates  the effectiveness of this method across various error types, as well as the pivotal role of alignment in enhancing performance.

Broadly speaking, the overcorrection challenge falls within the realm of uncontrollability of generative language models. Besides straightforward efforts to acquire more high-quality training data or employ specific pre-training strategies such as BART, this study introduces an alignment-based method that has demonstrated effectiveness in addressing this issue. Despite the improvement of our approach for decoder-only LLMs, their performance in CGEC still lags behind that of the strongest Seq2Seq models, even though they are smaller in size, which contradicts their outstanding performance in other NLP tasks. In future research, we will further exploring enhancing the performance of decoder-only LLMs for CGEC.

\section*{Acknowledgements}
We appreciate the anonymous reviewers for their valuable comments. This work was supported by the National Natural Science Foundation of China (No. 62176270) and the Guangdong Basic and Applied Basic Research Foundation (No. 2023A1515012832).

\section*{Limitations}
The potential limitations of our work are threefold. First, we have exclusively validated our approach on Chinese GEC datasets. However, our approach is language-independent, and it can be investigated in other languages. Second, our approach incurs additional training costs, as training alignment models and performing knowledge distillation are required. Third, our experiments are confined to 7B-scale LLMs using the QLoRA efficient fine-tuning technique. Due to computational resource constraints, we have not explored the impact of larger-scale LLMs and full-parameter fine-tuning, which may lead to improved performance.

\section*{Ethics Statement}
This work aims to propose a technical method to mitigate overcorrection caused by Seq2Seq models and decoder-only LLMs in Chinese grammatical error correction, which will not cause ethical issues. All datasets and models used in this work are publicly available, and we adhere strictly to their usage policies. We are committed to conducting our research in an ethical and responsible manner.

\bibliography{custom}
\bibliographystyle{acl_natbib}

\clearpage

\appendix

\section{Instruction Templates}
\label{appendix:templates}
In our experiments, we explored various instruction templates and observed that the choice of instruction templates has a limited impact on the experimental results, particularly when the amount of training data is sufficient. Table~\ref{tab:instruction} presents the instruction templates $\mathcal{T}_{\text{gec}}$ and $\mathcal{T}_{\text{align}}$ used in our experiments for tuning LLMs. The instruction template comprises an input field that provides the source text and a response field that denotes the target text.
\begin{CJK*}{UTF8}{gbsn}
\begin{table}[h]
\centering
    % \vspace{-2mm}
    \resizebox{\columnwidth}{!}{
    \begin{tabular}{l|l}
    \toprule
    \textbf{LLM} & \textbf{Instruction for correction model $\mathcal{T}_{\text{gec}}$} \\
    \hline
    \\[-1em]
    Baichuan2   & 纠正输入句子中的语法错误，并输出 \\
                & 正确的句子。\\
                & (Trans.: \textit{Correct grammatical errors}\\
                & \textit{in the input sentence and output the}\\
                & \textit{correct sentence.})\\
                & Input: \{Source\}  \\
                & Response: \{Target/Output\} \\
    \hline
    \\[-1em]
    Chinese-LLaMA2      &  \#\#\# Instruction: 纠正输入句子中的语 \\
                        & 法错误，并输出正确的句子。\\
                        & (Trans.: \textit{Correct grammatical errors}\\
                        & \textit{in the input sentence and output the}\\
                        & \textit{correct sentence.})\\
                        & \#\#\# Input: \{Source\}   \\
                        & \#\#\# Response: \{Target/Output\}  \\
    \hline
    \hline
    \\[-1em]
    \textbf{LLM} & \textbf{Instruction for alignment model $\mathcal{T}_{\text{align}}$} \\
    \hline
    \\[-1em]
    Baichuan2    & 对齐输入中用``\textbackslash t''分隔的两个句子，并输\\
                 & 出没有语法错误的句子。 \\
                 & (Trans.: \textit{Align the two sentences separated}\\
                 & \textit{by ``\textbackslash t'' in the input and output the sentence}\\
                 & \textit{without grammatical errors.})\\
                 & Input: \{Source\} \textbackslash t \{Initial Correction\}  \\
                 & Response: \{Target/Output\} \\
    \hline
    \\[-1em]
    Chinese-LLaMA2      & \#\#\# Instruction: 对齐输入中用``\textbackslash t''分隔的两 \\
                        & 个句子，并输出没有语法错误的句子。 \\
                        & (Trans.: \textit{Align the two sentences separated}\\
                        & \textit{by ``\textbackslash t'' in the input and output the sentence}\\
                        & \textit{without grammatical errors.})\\
                        & \#\#\# Input: \{Source\} \textbackslash t \{Initial Correction\}   \\
                        & \#\#\# Response: \{Target/Output\} \\
    
    \bottomrule
    \end{tabular}
    }
    \caption{Instruction templates for the correction model and alignment models, where ``Trans.'' denotes the translation of the instruction.} 
    \label{tab:instruction}
    % \vspace{-5mm}
    
\end{table}
\end{CJK*}

\section{Datasets}
\label{appendix:datasets}
The statistics of the datasets used in our experiments are shown in Table~\ref{tab:dataset_statistics}. For CSL learner data, we adopted the same training set as \citet{zhang-etal-2022-mucgec}, which involves discarding all samples without grammatical errors in the Lang8 and HSK datasets and replicating the HSK dataset five times and combining with the Lang8 dataset, resulting in a total of 1,568,885 sentence pairs.
\begin{table}[h]
    \centering
    \resizebox{\columnwidth}{!}{
        \begin{tabular}{l c c c}
            \toprule
            \textbf{Train}  & \textbf{Source}  & \textbf{\#Sent}  &  \textbf{\#Error}   \\
            \hline
            \textbf{Lang8}  & Learner  & 1,220,906    & 1,092,285 (89.5\%) \\
            \textbf{HSK} & Learner  & 15,6870    & 95,320 (60.8\%) \\
            \textbf{FCGEC}   & Native  & 35,341    & 19,183 (54.3\%) \\
            \hline
            \hline
            \textbf{Dev}  & \textbf{Source}  & \textbf{\#Sent}  &  \textbf{\#Error}   \\
            \hline
            \textbf{MuCGEC-Dev}  & Learner  & 2,467    & 2,409 (97.6\%) \\
            \textbf{FCGEC} & Native  & 1,000    & 563 (56.3\%) \\
            \hline
            \hline
            \textbf{Test}  & \textbf{Source}  & \textbf{\#Sent}  &  \textbf{\#Error}   \\
            \hline
            \textbf{NLPCC18-Test}  & Learner  & 2,000    & 1,983 (99.2\%) \\
            \textbf{FCGEC-Dev} & Native  & 2,000    & 1,101 (55.1\%) \\
            \textbf{NaCGEC-Test} & Native  & 5,869    & 5,612 (95.6\%) \\
            \bottomrule
        
        \end{tabular}
    }
    \caption{Statistics of the used CGEC datasets. \textbf{\#Sent} denotes the number of the sentences and \textbf{\#Error} denotes the number (the percentage) of the erroneous sentences.}
    \label{tab:dataset_statistics}
    % \vspace{-3mm}
\end{table}

\section{Experimental Details}

\subsection{Training Details}
\label{appendix:training_details}
\paragraph{Training on Native Speaker Datasets}
Since FCGEC contains only 35,341 training samples, which is insufficient for model training, we performed continuous training on the FCGEC training set with the model trained on the CSL learner data.
\paragraph{Training of Alignment Models}
As described in Section~\ref{method:alignment_model}, before training the alignment models, we need to obtain initial corrections using an initial correction model. For this purpose, we divided the training data into two parts, one for training the initial correction model and the other for training the alignment models. In the case of CSL learner data, 80\%\footnote{We experimented with different ratios, including 4:6, 5:5, and 8:2, and found that 8:2 works the best.} of the training samples are randomly selected to train the initial correction model. Then, this correction model is used to generate initial corrections for the remaining training samples. These initial corrections along with their corresponding source and target sentences are used to train the alignment models. For the native speaker datasets, we used the correction model trained on the CSL learner data to generate initial corrections on the FCGEC training set. 
\paragraph{Training of Alirector}
As outlined in Section~\ref{method:kd}, we perform knowledge distillation using the correction model as the student and the alignment models as the teachers. For this training, we used the same training set as that used for training the teachers. The student was initialized with the weights of the well-trained initial correction model.

\begin{table*}[t]
\centering
    \resizebox{0.9\textwidth}{!}{
    \begin{tabular}{lcccc}
        \toprule
        \textbf{Hyperparameter}      & \multicolumn{2}{c}{\textbf{NLPCC18}} & \multicolumn{2}{c}{\textbf{FCGEC/NaCGEC}}  \\
        \hline
        \\[-1em]
        \multicolumn{5}{c}{\textbf{Seq2Seq}} \\
        \hline
        \\[-1em]
        Backbone & Transformer-large & BART-large & Transformer-large & BART-large \\
        Batch size      & 1024 & 1024 & 256 & 256   \\
        Max Epochs      & 20 & 10 & 20 & 10 \\
        Max Length      & \multicolumn{4}{c}{128 (Source); 128 (Target)}   \\
        Learning Rate   & $3 \times 10^{-4}$ & $3 \times 10^{-5}$ & $3 \times 10^{-5}$ & $3 \times 10^{-5}$   \\
        Warmup Steps    & 3000 & 1000 & 100 & 100  \\
        Dropout         & 0.3 & 0.1 & 0.3 & 0.1      \\ 
        Dropout-Src     & 0.2 & 0.2 & 0.2 & 0.2   \\
        $\alpha$     & 0.7 & 0.5 & 0.5 & 0.9   \\
        $\beta$      & 1.0 & 1.5 & 0.5 & 0.5   \\
        $\tau$       & 1 & 1 & 1 & 1 \\
        Beam Size       & 10  & 10 & 10 & 10   \\
        \hline
        \hline
        \\[-1em]
        \multicolumn{5}{c}{\textbf{LLMs}} \\
        \hline
        \\[-1em]
        Backbone & Baichuan2-7B & Chinese-LLaMA2-7B & Baichuan2-7B & Chinese-LLaMA2-7B \\
        Batch size      & 1024 & 1024 & 256 & 256   \\
        Max Epochs      & 3 & 5 & 10 & 10 \\
        Max Length      & \multicolumn{4}{c}{192 (GEC); 256 (Alignment)}   \\
        Learning Rate   & $3 \times 10^{-5}$ & $3 \times 10^{-4}$ & $3 \times 10^{-5}$ & $3 \times 10^{-5}$   \\
        Warmup Steps    & 1000 & 1000 & 100 & 100  \\
        LoRA    & \multicolumn{4}{c}{target modules = all linears; lora rank = 8; lora alpha = 16, lora dropout = 0.05} \\
        $\alpha$     & 0.3 & 0.5 & 0.3 & 0.5   \\
        $\beta$      & 1.5 & 2.0 & 0.5 & 1.0   \\
        $\tau$       & 1 & 1 & 1 & 1 \\
        Beam Size       & 10 & 10 & 10 & 10   \\
        \bottomrule
    \end{tabular}
    }
    \caption{Hyperparameter settings in our experiments.}
    \label{tab:hyperparameter}
\end{table*}
\subsection{Implementation Details}
\label{appendix:implementation_details}
For Seq2Seq model training, following \citet{zhang-etal-2022-syngec}, we utilized the Dropout-Src technique \citep{junczys-dowmunt-etal-2018-approaching} that applies dropout on input embeddings for alleviating over-fitting. As for LLMs tuning, considering the time and computational resources, we applied QLoRA \citep{dettmers2023qlora} for efficient fine-tuning instead of full-parameter fine-tuning. Our code implementation mainly follows the \textit{Alpaca LoRA} project\footnote{\url{https://github.com/tloen/alpaca-lora}}, and is based on the Huggingface Transformers \citep{wolf-etal-2020-transformers} and bitsandbytes\footnote{\url{https://github.com/TimDettmers/bitsandbytes}} \citep{dettmers2022llmint8} toolkit in Pytorch.~We searched for the optimal value of $\alpha$ in \{0.1, 0.3, 0.5, 0.7, 0.9\}, $\beta$ in \{0.5, 1.0, 1.5, 2.0\} and the temperature $\tau$ in \{1, 2, 3, 4, 5\} on the development set. We used the Adam optimizer \citep{kingma2014adam} and polynomial learning rate decay. The hyperparameter settings are presented in Table~\ref{tab:hyperparameter}. All experiments are carried out on 8 GeForce RTX 4090 24GB GPUs.

\begin{table}[t]

    \begin{adjustbox}{width=\columnwidth,center}
    \begin{tabular}{l|ccc ccc}
        \toprule
        \multirow{2}{*}{\textbf{Method}} & \multicolumn{3}{c}{\textbf{NaCGEC-Test}} 
        & \multicolumn{3}{c}{\textbf{FCGEC-Dev}}  \\
        & \textbf{P} & \textbf{R} & $\textbf{F}_{0.5}$    & \textbf{P} & \textbf{R} & $\textbf{F}_{0.5}$ \\ 
        \hline 
        \textbf{Vanilla FT}  & 62.93 & 44.50 & 58.12   & 56.26 & 40.71 & 52.27   \\
        \textbf{predict-and-align}  & \textbf{67.21} & 45.61 & \textbf{61.39}  & \textbf{62.60} & 37.43 & \textbf{55.18}   \\
        \ \textit{\, repl.} $\text{src+src}$   & 66.05 & 41.78 & 59.18   & 61.50 & 32.25 & 52.06   \\
        \ \textit{\, repl.} $\text{pred+pred}$  & 59.98 & \textbf{49.56} & 57.53       & 50.59 & \textbf{42.48} & 48.73   \\
        \hline
        \textbf{Alirector} & 66.04 & 45.91 & 60.71  & 58.55 & 39.74 & 53.49   \\
        \bottomrule
    \end{tabular}
    \end{adjustbox}
    \caption{Results of the potential of alignment on FCGEC-Dev, where "\textit{repl.}" is short for "replace".}
    
    \label{tab:alignment_results}
% \vspace{-3mm}
\end{table}
\section{Potential of Alignment}
\label{appendix:alignment_model}
The alignment models are introduced to mitigate overcorrection by leveraging the alignment information between the source sentence and the initial correction. To demonstrate the potential of the alignment models, we employed a BART-based alignment model to align the predictions of a Baichuan2-based correction model, and present the comparison results between vanilla fine-tuning (i.e., without alignment) and predict-and-align method in Table~\ref{tab:alignment_results}. From the results, we note that predict-and-align improves precision and $F_{0.5}$ by a large margin compared to vanilla fine-tuning. Notably, predict-and-align even outperforms our Alirector, highlighting the effectiveness and the potential of the two-stage alignment method. Moreover, when only source information is retained in the input (namely \textit{repl.} src+src), we observe high precision but low recall, while retaining only prediction information (namely \textit{repl.} pred+pred) exhibits the opposite trend. This observation once again emphasizes the role of alignment in reducing both overcorrection and undercorrection.

\section{Additional Experimental Results}
\label{appendix:more_exp_results}

\subsection{Results on FCGEC-Test}
We conducted additional experiments on FCGEC-Test\footnote{FCGEC-Test provides online evaluation at \url{https://codalab.lisn.upsaclay.fr/competitions/8020}.}, the test set of FCGEC \citep{xu-etal-2022-fcgec}, for more comprehensive evaluation. As shown in Table~\ref{tab:fcgec_test_results}, our Alirector improves the P/$F_{0.5}$ score by 5.59/1.77 over vanilla fine-tuning on BART, while the improvement is 4.37/2.25 on Baichuan2-7B. In contrast, the Copy method has only a minor improvement over vanilla fine-tuning.
\definecolor{mygray}{gray}{.9}
\definecolor{ggreen}{rgb}{0.0, 0.6, 0.0}

\begin{table}[h!]
    \begin{adjustbox}{width=\columnwidth,center}
        \begin{tabular}{l c ccc}
            \toprule
            \multirow{2}{*}{\textbf{Model}}      & \multirow{2}{*}{\textbf{Method}}           & \multicolumn{3}{c}{\textbf{FCGEC-Test}} \\
                &        & \textbf{P} & \textbf{R} & $\textbf{F}_{0.5}$ \\ 
            \hline
            \\[-0.9em]
            \multirow{3}{*}{BART} & Vanilla FT & 63.85  & \textbf{40.16}  & 57.11  \\
            & Copy         & 65.31  & 39.45  & 57.74    \\
            & Alirector      & \textbf{69.44}  & 36.60  & \textbf{58.88}  \\
            \hline 
            \\[-0.9em]
            \multirow{3}{*}{Baichuan2-7B}     & Vanilla FT    & 60.12  & \textbf{37.21}  & 53.53  \\
            & Copy         & 62.14  & 35.47  & 54.01     \\
            & Alirector      & \textbf{64.49}   & 36.22  & \textbf{55.78} \\             
            \bottomrule
        \end{tabular}
    \end{adjustbox}
    \caption{Results on FCGEC-Test.}
    
    \label{tab:fcgec_test_results}
    % \vspace{-3mm}
\end{table}

\vspace{-0.3cm}
\subsection{Results on More LLMs}
We also implemented our Alirector method on other Chinese LLMs, namely Yi-6B\footnote{\url{https://huggingface.co/01-ai/Yi-6B}}, a Chinese LLM that employs the same architecture as LLaMA \citep{llama}, and ChatGLM3-6B\footnote{\url{https://huggingface.co/THUDM/chatglm3-6b-base}}, a chat model based on GLM \citep{du-etal-2022-glm}. The results on NLPCC18-Test are shown in Table~\ref{tab:more_llms_results}. We observe that our Alirector method improves the precision/$F_{0.5}$ score over vanilla fine-tuning by 3.62/1.75 on Yi-6B and 3.14/1.05 on ChatGLM3-6B, respectively. This highlights the generalizability of Alirector across other LLMs.
However, the performance of Yi-6B and ChatGLM3-6B lags significantly behind that of Baichuan2-7B. This discrepancy can be attributed to the different capabilities achieved through pre-training.
\definecolor{mygray}{gray}{.9}
\definecolor{ggreen}{rgb}{0.0, 0.6, 0.0}

\begin{table}[t]
    \begin{adjustbox}{width=\columnwidth,center}
        \begin{tabular}{l c ccc ccc}
            \toprule
            \multirow{2}{*}{\textbf{Model}}      & \multirow{2}{*}{\textbf{Method}}           & \multicolumn{3}{c}{\textbf{NLPCC18-Test}} \\
                &        & \textbf{P} & \textbf{R} & $\textbf{F}_{0.5}$ \\ 
            \hline
            \\[-0.9em]
            \multirow{3}{*}{Yi-6B} & Vanilla FT & 50.61  & \textbf{25.22}  & 42.13  \\
            & Copy         & 50.11  & 25.01  & 41.73    \\
            & Alirector      & \textbf{54.23}  & 24.89  & \textbf{43.88}  \\
            \hline 
            \\[-0.9em]
            \multirow{3}{*}{ChatGLM3-6B}     & Vanilla FT    & 48.61  & \textbf{26.46}  & 41.64  \\
            & Copy         & 49.64 &25.23 &41.59       \\
            & Alirector      & \textbf{51.75}   & 25.11  & \textbf{42.69} \\             
            \bottomrule
        \end{tabular}
    \end{adjustbox}
    \caption{Results on Yi-6B and ChatGLM3-6B.}
    
    \label{tab:more_llms_results}
    \vspace{-3mm}
\end{table}

\section{Case Study}
\label{appendix:case_study}
We provide four examples in Table~\ref{tab:example} to illustrate the effectiveness of our Alirector in mitigating overcorrection. We can note that the vanilla fine-tuned model often tends to overcorrect by modifying the error-free characters. In contrast, Alirector is able to correct all the errors in the sentence while preserving the error-free characters. These cases intuitively show that Alirector learns to identify and correct the potential errors in the sentence while actively avoiding overcorrection as much as possible.
\begin{CJK*}{UTF8}{gbsn}
\begin{table*}[ht]
    \begin{adjustbox}{width=\textwidth,center}
        \begin{tabular}{c|l}
            \toprule
            \multirow{2}{*}{\textbf{Source}} &  在过去一年，我\textcolor{orange}{校}采取了一系列卓有成效的\textcolor{orange}{改进}方法。  \\
            & In the past year, our school has adopted a series of effective improvement methods.  \\
            \hline

            \\[-0.5em]
            \multirow{2}{*}{\textbf{Target}} & 在过去一年，我校采取了一系列卓有成效的改进措施。  \\
            & In the past year, our school has adopted a series of effective improvement measures.  \\
            \hline
            
            \\[-0.5em]
            \multirow{2}{*}{\textbf{Vanilla FT}} & 在过去一年，我\textcolor{red}{们}采取了一系列卓有成效的\textcolor{red}{解决}\textcolor{blue}{措施}。 (\textcolor{red}{\XSolidBrush}) \\
            & In the past year, we have adopted a series of effective solution measures.  \\
            \hline
            
            \\[-0.5em]
            \multirow{2}{*}{\textbf{Alirector}} & 在过去一年，我\textcolor{orange}{校}采取了一系列卓有成效的\textcolor{orange}{改进}\textcolor{blue}{措施}。 (\textcolor{blue}{\Checkmark})  \\
            & In the past year, our school has adopted a series of effective improvement measures.  \\
            
            \hline
            \hline

            \\[-0.5em]
            \multirow{3}{*}{\textbf{Source}} & 他已经回国，现就任于北大医学部\textcolor{orange}{体育教授}。  \\
            & He has returned to China and is now serving in a professor of physical education in the Peking University \\
            & Health Science Center.  \\ 
            \hline

            \\[-0.5em]
            \multirow{3}{*}{\textbf{Target}} & 他已经回国，现就任于北大医学部，担任体育教授。  \\
            & He has returned to China and is now serving as a professor of physical education in the Peking University \\
            & Health Science Center. \\
            \hline
            
            \\[-0.5em]
            \multirow{2}{*}{\textbf{Vanilla FT}} & 他已经回国，现就任于北大医学部\textcolor{red}{\sout{体育教授}}。 (\textcolor{red}{\XSolidBrush})\\
            & He has returned to China and is now serving in the Peking University Health Science Center.  \\
            \hline
            
            \\[-0.5em]
            \multirow{3}{*}{\textbf{Alirector}} & 他已经回国，现就任于北大医学部，\textcolor{blue}{担任}\textcolor{orange}{体育教授}。 (\textcolor{blue}{\Checkmark})  \\
            & He has returned to China and is now serving as a professor of physical education in the Peking University \\
            & Health Science Center. \\

            \hline
            \hline

            \\[-0.5em]
            \multirow{2}{*}{\textbf{Source}} & 斯诺登\textcolor{orange}{虽然决定}了自己的住处，而且出于安全考虑他不会公布住址。  \\
            & Snowden has decided where he is going to live, though, and he will not release his address for security reasons.  \\ 
            \hline

            \\[-0.5em]
            \multirow{2}{*}{\textbf{Target}} & 斯诺登虽然决定了自己的住处， 但是出于安全考虑他不会公布住址。  \\
            & Snowden has decided where he will live, but he will not release his address for security reasons. \\
            \hline
            
            \\[-0.5em]
            \multirow{2}{*}{\textbf{Vanilla FT}} & \textcolor{red}{虽然}斯诺登\textcolor{red}{决定}了自己的住处，\textcolor{blue}{但是}出于安全考虑他不会公布住址。 (\textcolor{red}{\XSolidBrush})  \\
            & Although Snowden has decided where he will live, he will not release his address for security reasons. \\
            \hline
            
            \\[-0.5em]
            \multirow{2}{*}{\textbf{Alirector}} & 斯诺登\textcolor{orange}{虽然决定}了自己的住处，\textcolor{blue}{但是}出于安全考虑他不会公布住址。 (\textcolor{blue}{\Checkmark})   \\
            & Snowden has decided where he will live, but he will not release his address for security reasons. \\

            \hline
            \hline
            
            \\[-0.5em]
            \multirow{3}{*}{\textbf{Source}} &  这样不仅有助于维护国家安全和社会稳定，而且\textcolor{orange}{有利于提高工作效率}，有利于金融机构落实存款实名制。  \\
            & This will not only help maintain national security and social stability, but also help improve work efficiency and  \\ 
            & help financial institutions implement the real-name deposit system.  \\
            \hline

            \\[-0.5em]
            \multirow{3}{*}{\textbf{Target}} & 这样不仅有利于提高工作效率，有利于金融机构落实存款实名制，而且有助于维护国家安全和社会稳定。  \\
            & This will not only help improve work efficiency, but also help financial institutions implement the real-name  \\
            & deposit system and help maintain national security and social stability.  \\
            \hline
            
            \\[-0.5em]
            \multirow{3}{*}{\textbf{Vanilla FT}} & 这样不仅\textcolor{red}{\sout{有利于提高工作效率}}，有利于金融机构落实存款实名制，而且\textcolor{blue}{有助于维护国家安全和社会稳定}。 (\textcolor{red}{\XSolidBrush})  \\
            & This will not only help financial institutions implement the real-name deposit system, but also help maintain  \\
            & national security and social stability.  \\
            \hline
            
            \\[-0.5em]
            \multirow{3}{*}{\textbf{Alirector}} & 这样不仅\textcolor{orange}{有利于提高工作效率}，有利于金融机构落实存款实名制，而且\textcolor{blue}{有助于维护国家安全和社会稳定}。 (\textcolor{blue}{\Checkmark})   \\
            & This will not only help improve work efficiency, but also help financial institutions implement the real-name  \\
            & deposit system and help maintain national security and social stability.  \\

            \bottomrule
        \end{tabular}
    \end{adjustbox}
    \caption{A case study of vanilla fine-tuning and our Alirector using Baichuan2-7B on FCGEC-dev and NaCGEC-Test, where overcorrected characters and their error-free counterparts are highlighted in red and orange, respectively, and correct edits are highlighted in blue.}
    \label{tab:example}
    % \vspace{-3mm}
\end{table*}
\end{CJK*}

\section{Explanation of Term "Alignment"}
\label{appendix:explanation_of_alignment}
We adopt the term "alignment" in this paper as it works in a similar way to the alignment in RLHF \citep{rlhf}. RLHF aims to align the LLM's response with human preferences, while our alignment model seeks to refine the GEC model's over-corrected predictions by aligning them with the source text. From this perspective, our approach ensures that the corrections are more accurate and contextually appropriate, maintaining faithfulness to the original input while avoiding unnecessary changes. Additionally, \citet{ji2024aligner} use an \textit{Aligner} to refine the initial output of LLMs, which shares a similar concept with our paper.

\end{document}